%% file: main.tex
\documentclass[10pt]{article}
\usepackage[accepted]{tmlr}

\usepackage{microtype}
\usepackage{graphicx}
\usepackage{hyperref}
\usepackage{url}
\usepackage{booktabs}
\usepackage{multirow}
\usepackage{amsmath}
\usepackage{amssymb}
\usepackage{tikz}
\usepackage{placeins}
\usepackage{float}
\usepackage{threeparttable}
\usetikzlibrary{arrows.meta,positioning,calc}

\input{math_commands.tex}

\definecolor{boxblue}{RGB}{0,114,178}
\definecolor{pastein}{RGB}{213,94,0}
\definecolor{timegray}{RGB}{80,80,80}
\definecolor{darkblue}{rgb}{0, 0, 0.5}
\hypersetup{colorlinks=true, citecolor=darkblue, linkcolor=darkblue, urlcolor=darkblue}

\title{Noise Your Prompt: Noising Conditioning Tokens in Continuous Diffusion Language Models}

\author{\name Justin Jung \email justin@lateralintelligence.org \\
      \addr Lateral Intelligence}

\def\month{09}  
\def\year{2026} 
\def\openreview{\url{https://openreview.net/forum?id=carxDjFbtO}} 

\begin{document}

\maketitle

\begin{abstract}
  We revisit a standard accepted practice in the continuous diffusion language model
  literature of fixing conditioning prompt tokens clean during training.
  We make a very simple modification: also noise the conditioning prompt tokens during training.
  We demonstrate that under this modified training objective, we achieve better generalization
  in combinatorial reasoning tasks such as Sudoku and N-Queens, with the largest gains on harder variants
  ($3.73\% \to 24.65\%$ solve rate on Sudoku Hard), and increased diversity of generated solutions ($50.60\% \to 73.79\%$ coverage on
  10×10 N-Queens). We also show measurable improvements to natural language generation quality
  in modest dataset regimes with Gigaword summarization, but notably demonstrate that gains do not
  transfer to all natural language tasks (e.g open ended dialogue generation).
  Our method is a single line change to the training objective, requires no additional inference costs by default,
  and provides the flexibility of classifier-free guidance inspired guided sampling. Our  
  \href{https://github.com/LateralIntelligence/noise-your-prompt} {code} is publicly available.

\end{abstract}

\section{Introduction}
\label{sec:intro}
Fundamentally, diffusion and flow models learn a score or velocity corresponding to an ODE
which can be integrated to transform a base distribution to the target data distribution.
Their objective can be parameterized in different ways,
including the denoising parametrization $\hat{x}_1(x_t)$, where the goal is to predict a
clean target data $x_1$ given a current noisy input $x_t$.

Our framework is to view this denoising learning objective from the lens of conditional prediction:
given a noisy input, predict a clean target. We then see the varying levels of noise
corruption of the input as a spectrum of prediction task difficulties, where the model is trained on
harder (more noisy) inputs and easier (less noisy) inputs.

In practice for language modelling, conditional generation is the main paradigm: given a
conditioning prompt, generate a response. Currently, conditional generation in continuous
diffusion language models and flow language models utilize the following learning objective: given
un-noised conditioning tokens $x^c$ and a noised response $y_t$, predict the target response $y_1$.
Likewise in sampling, the conditioning tokens are fixed and not noised. 

We improve this conditional learning objective
by allowing conditional tokens to also be noised, where the model is given
a noised conditional input $x_t^c$ along with the current noised response $y_t$. We see this modification
as extending the range of prediction tasks difficulties, where the prediction problem $(x, y_t) \to y$ is made more difficult not
just by increasing the amount of noise in the current response input $y_t$ but also by
increasing the amount of noise in the conditioning input. Moreover, by noising the conditioning input,
our model has a wider target that it can learn from
(also needs to predict the denoised conditioning input), making learning more
efficient. Varying the level of noise
on the conditioning input can be seen as conditioning augmentation.
Empirically, we evaluate on various reasoning and generalization tasks, specifically Sudoku and N-Queens,
and natural language generation tasks of summarization and open ended dialogue generation.

\paragraph{Scope and contributions.}
Our experimental results utilize flow language models
on two combinatorial constraint satisfaction benchmarks, Sudoku and N-Queens, and on 
two natural language generation datasets, Gigaword summarization and open ended dialogue generation.  
We provide evidence for three claims: (i) Noising conditioning information during training 
yields improved solve rates on unseen puzzles versus the default paste-in setup, with the largest
gains on the more difficult puzzle settings (Sections~\ref{sec:sudoku} and~\ref{sec:nqueens}).
(ii) Noised prompt training produces more diverse generated solutions, as measured 
by solution coverage on N-Queens (Section ~\ref{sec:nqueens}). (iii) Noised prompt training improves 
summarization generation quality in the modest dataset regime. With natural language generation, we demonstrate 
measurable improvements in Gigaword summarization but find null results in open dialogue generation, highlighting 
that noised prompt training can lead to improvements in some but not all natural language tasks.
The takeaway is that the default setup of keeping conditioning tokens clean during
training of diffusion and flow language models should be revisited; a trivial change to also 
noise conditioning tokens during training improves both accuracy and diversity in our controlled 
puzzle testbeds and shows measurable improvements to generation quality in certain natural language tasks.

\begin{figure}[t]
\centering
\includegraphics[width=\linewidth]{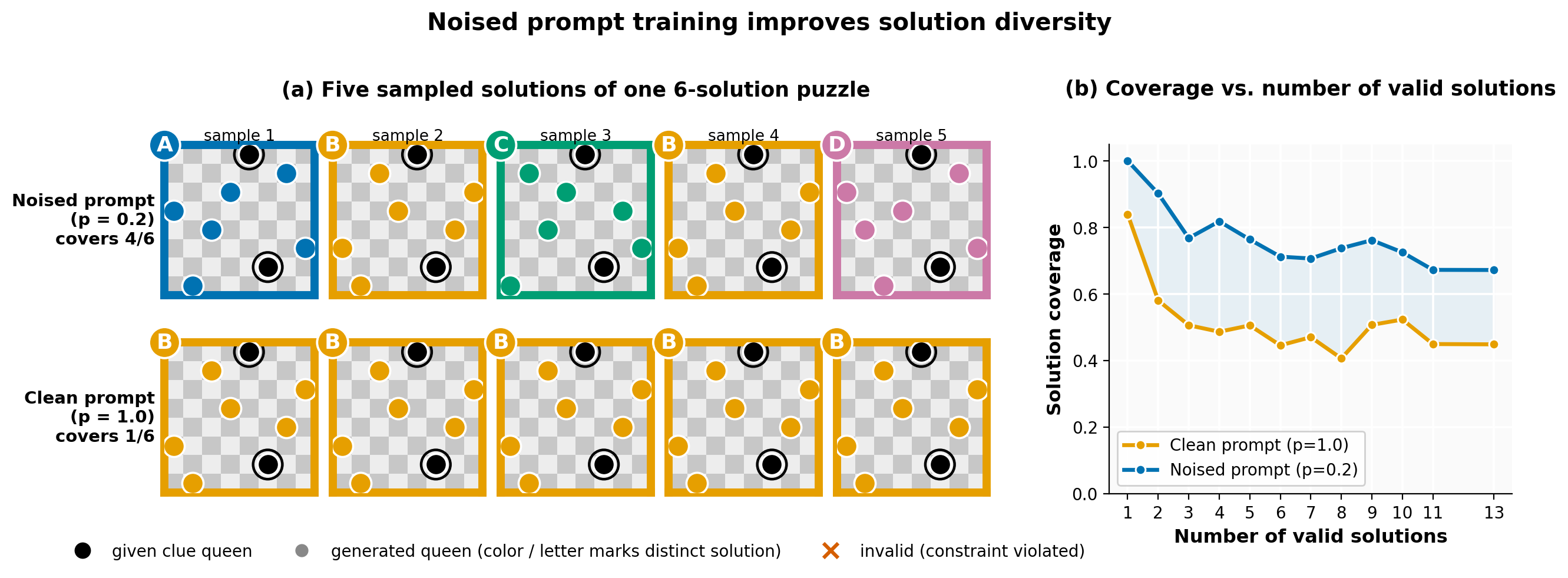}
\caption{\textbf{Noising the conditioning tokens improves solution diversity.} \emph{(a)} Five board generations of a single held-out N-Queens ($8\times8$)
puzzle that admits six distinct valid solutions (distinguished by color and letter). A model trained with noised conditioning tokens ($p_\text{condclean}{=}0.2$, top) spreads its
generations across the solution set ($4/6$ distinct solutions), whereas the standard ``paste-in''
model that keeps conditioning tokens clean throughout training ($p_\text{condclean}{=}1.0$, bottom) mode-collapses
onto a single completion. \emph{(b)} Solution coverage as a function of the number of valid
solutions admitted by the input board, over $750$ held-out $10\times10$ puzzles with $k{=}20$
generations each. The noised-prompt model dominates the clean-prompt model at every
solution count. We quantify this gap in Section~\ref{sec:nqueens}.}
\label{fig:teaser}
\end{figure}

\section{Background and Related Work}
\label{sec:background}

\paragraph{Diffusion and flow generative models.}
Denoising diffusion and score-based models generate data by gradually corrupting samples toward a
simple base distribution and learning to reverse this process~\citep{ddpm, song2021sde}. Flow
matching and stochastic interpolants~\citep{lipman2023flow, albergo2023interpolants} give a closely
related and more general view: a time-dependent interpolation between the base and data distributions
defines a velocity field whose ordinary differential equation (ODE) transports noise to data, with
diffusion as a special case. The learning target admits several equivalent
parameterizations---predicting the injected noise, the velocity, the score, or directly the clean
datapoint $x_1$ from a noised input $x_t$ ($x_1$-prediction). We work throughout in the
flow and $x_1$-prediction setting.

\paragraph{Diffusion and flow language models.}
Applying these models to text is complicated by the discrete nature of tokens. One line of work operates
directly in discrete state spaces, defining absorbing or uniform corruption processes over tokens:
D3PM~\citep{d3pm}, score-entropy discrete diffusion (SEDD)~\citep{sedd}, masked diffusion language
models (MDLM)~\citep{mdlm}, MD4~\citep{shi2025simplifiedgeneralizedmaskeddiffusion}, and large-scale
masked diffusion LMs such as LLaDA~\citep{llada}.
A second line embeds tokens into a continuous space, running continuous gaussian diffusion over those
representations, and converting back to tokens: Diffusion-LM~\citep{diffusion_lm}, DiffuSeq~\citep{diffuseq},
and CDCD~\citep{cdcd}. Recent work operates specifically on one-hot vectors on the probability simplex or 
on learned / pre-trained embeddings: Flow Map Language
Models~\citep{flm}, Discrete Flow Maps~\citep{discrete_flow_maps}, hyperspherical
flows~\citep{hyperspherical_flows}, and embedded language flows (ELF)~\citep{elf}. Our work builds on
this continuous-flow family and adopts its standard denoising ($x_1$-prediction) objective. By default,
these models do conditional generation by holding conditioning tokens clean throughout training
and moreover during sampling, where conditioning tokens are re-pasted at every sampling step ~\citep{flm, elf}.
We term this default method as \emph{paste-in} conditioning.

Diffusion forcing~\citep{diffusion_forcing}, AR-Diffusion~\citep{ar_diffusion}, and concurrent work
TTCD~\citep{bansal2026tokentimecontinuousdiffusion} assign per-token noise levels rather
than a single shared noise level for the entire sequence. This framework
motivates the notion of multiple noise levels across the sequence, where conditioning tokens may
carry a different noise level than response tokens.

\paragraph{Classifier-free guidance.}
Classifier-free guidance~\citep{cfg} jointly trains a conditional and an unconditional model by
randomly dropping the conditioning signal, then combines their predictions at sampling time to trade
off fidelity against diversity. The amount of conditioning signal in the sequence is binary:
it is either fully injected or fully discarded. We use CFG as inspiration and allow the amount of 
conditioning signal to be smoothly interpolated by the amount of noise (parametrized by time $t \in [0,1]$). 
Moreover, we can analogously 
combine the model's predictions under a unmodified clean prompt and the model's predictions 
under a noised prompt during inference for guided sampling. 

\paragraph{Corrupting conditioning tokens.}
Corrupting the conditioning information during training has appeared in several
settings, though rarely for language. In image generation, cascaded diffusion
models~\citep{cascaded} apply conditioning augmentation, adding noise to the
low-resolution conditioning image fed to each super-resolution stage; this improves robustness to previously
produced artifacts. Classifier-free guidance~\citep{cfg} can likewise be seen as the extreme case of corrupting the conditioning by
removing it entirely. For language, corrupting the conditioning is far less explored and the evidence
is mixed. Blockwise SFT for diffusion language models~\citep{blockwise_sft} reports that model performance
degrades as the amount of noise on noisy prefixes increases,
and so explicitly keeps the prefix clean.

Most directly related, prompt infilling work for masked diffusion language
models~\citep{prompt_infill} unlocks prompt infilling by masking the prompt as well as the response
during training, rather than masking the response alone. ~\citep{prompt_infill} generate 
multiple prompts and select the best one, offering an alternative to prompt engineering.
In contrast, our method noises conditioning tokens in continuous diffusion models during training 
across a continuum of corruption levels, which can be seen as expanding the range of prediction tasks.
Moreover, our method does not require additional inference costs and by default requires no changes to inference.

\section{Method}
\label{sec:method}

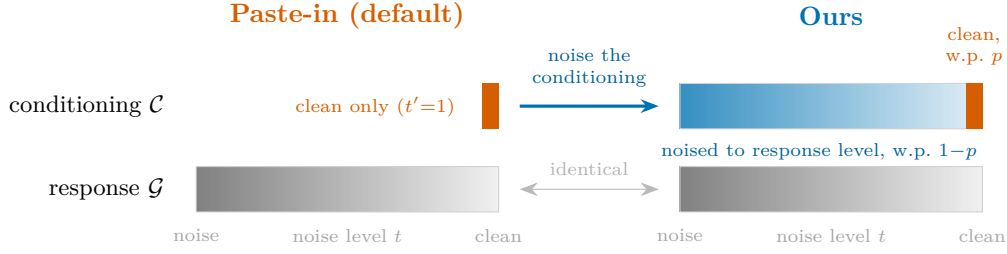
\begin{figure}[t]
  \centering
  \begin{tikzpicture}[font=\small, >={Stealth[length=2.4mm]}]
  \def\W{4.0}\def\h{0.30}
  \def\yc{1.25}\def\yr{0.15}
  \def\Lx{0}\def\Rx{6.4}
  
  
  \fill[pastein] (\Lx+\W-0.22,\yc-\h) rectangle (\Lx+\W,\yc+\h);
  \node[pastein,font=\scriptsize,anchor=east] at (\Lx+\W-0.45,\yc) {clean only ($t'{=}1$)};
  \shade[left color=black!50,right color=black!6] (\Lx,\yr-\h) rectangle (\Lx+\W,\yr+\h);
  \draw[gray!35] (\Lx,\yr-\h) rectangle (\Lx+\W,\yr+\h);
  
  \shade[left color=boxblue!80,right color=boxblue!12] (\Rx,\yc-\h) rectangle (\Rx+\W,\yc+\h);
  \draw[gray!35] (\Rx,\yc-\h) rectangle (\Rx+\W,\yc+\h);
  \fill[pastein] (\Rx+\W-0.22,\yc-\h) rectangle (\Rx+\W,\yc+\h);
  \node[pastein,font=\scriptsize,anchor=south,align=center] at (\Rx+\W-0.11,\yc+\h+0.12) {clean,\\ w.p.\ $p$};
  \node[boxblue!85!black,font=\scriptsize,anchor=north] at (\Rx+\W*0.46,\yc-\h-0.06) {noised to response level, w.p.\ $1{-}p$};
  \shade[left color=black!50,right color=black!6] (\Rx,\yr-\h) rectangle (\Rx+\W,\yr+\h);
  \draw[gray!35] (\Rx,\yr-\h) rectangle (\Rx+\W,\yr+\h);
  
  \node[pastein,font=\bfseries] at (\Lx+\W/2,2.45) {Paste-in (default)};
  \node[boxblue,font=\bfseries] at (\Rx+\W/2,2.45) {Ours};
  \node[anchor=east] at (\Lx-0.3,\yc) {conditioning $\mathcal{C}$};
  \node[anchor=east] at (\Lx-0.3,\yr) {response $\mathcal{G}$};
  \foreach \ox in {\Lx,\Rx}{
    \node[anchor=north,font=\scriptsize,gray!70] at (\ox,\yr-\h-0.1) {noise};
    \node[anchor=north,font=\scriptsize,gray!70] at (\ox+\W,\yr-\h-0.1) {clean};
    \node[anchor=north,font=\scriptsize,gray!70] at (\ox+\W/2,\yr-\h-0.1) {noise level $t$};
  }
  
  \draw[boxblue,->,line width=1.2pt] (\Lx+\W+0.28,\yc) -- (\Rx-0.28,\yc);
  \node[boxblue!85!black,font=\scriptsize,align=center,anchor=south] at (5.2,\yc+0.14) {noise the\\ conditioning};
  \draw[gray!55,<->] (\Lx+\W+0.28,\yr) -- (\Rx-0.28,\yr);
  \node[gray!60,font=\scriptsize,anchor=south] at (5.2,\yr+0.06) {identical};
  \end{tikzpicture}
  \caption{\textbf{Visualization of our training objective.} Standard conditional training
  keeps conditioning tokens at fixed clean noise level ($t'{=}1$). In contrast, we train over a range of
conditioning noise levels (blue region), exposing the model to a range of conditioning
levels. At generation time we follow standard practice and integrate with the conditioning tokens held clean.}
  \label{fig:method}
  \end{figure}


Our method is simple and requires minimal modification from the default
training objective of flow language models. Simply put, with some probability $p_\text{condclean}$ we keep the conditioning tokens clean; else,
we noise the conditioning tokens as well.

During sampling, following the default practice in diffusion/flow language model literature ~\citep{flm, elf,hyperspherical_flows}, we keep the
conditioning tokens clean and paste-in clean tokens at each sampling step. To prevent a large
training and inference mismatch, with some probability $p_\text{condclean}$ we choose to keep the
conditioning tokens clean to supply as input. The model then predicts using inputs
$(x_{t=1},y_t)$, where $x_{t=1}$ are the clean conditioning tokens and $y_t$ are the noised
response tokens.

At the same time, we do not want the model to become too reliant on the conditioning tokens and
possibly learn a brittle mapping $x \to y$. Thus, with the remaining
probability $1- p_\text{condclean}$, we noise our conditioning tokens, reducing the amount of conditioning signal 
the model has access to. It receives as input
both noised conditioning tokens and noised response tokens $(x_t, y_t)$. As visualized in Figure~\ref{fig:method}, 
 by additionally noising our conditioning tokens, our model is trained under a
broader curriculum of conditional prediction tasks, where the model receives both a lot of 
conditioning signal (making the prediction task easier) and little conditioning signal
(making the prediction task harder).  

Additionally, we make use of the conditioning tokens in the loss and have the model 
predict denoised versions of the conditioning tokens $x_t$. This utilizes all the tokens in the denoising
objective, making the training objective more efficient (Section~\ref{sec:ablations}).

\subsection{Setup}
\label{sec:method-setup}
We work with an in-place \emph{infilling} layout. Each example is a single sequence of length $L$
(Sudoku: $L=81$; N-Queens: $L=n^2$) over a small vocabulary. The model receives a sequence
of per token noise levels; the model receives no formal distinction
between conditioning tokens $\mathcal{C}$ and response tokens $\mathcal{G}$. We compute   
the loss over the full sequence.

\subsection{Corruption process and varying conditioning signal}
\label{sec:method-objective}
We adopt a continuous-time flow ($x_1$-prediction) model with per-token noise levels
(as in \emph{diffusion forcing}~\citep{diffusion_forcing}): each position $i$ carries its own time
$\tau_i \in [0,1]$, with $\tau_i = 1$ denoting a clean token and $\tau_i = 0$ full noise, and
$x_i(\tau_i)$ is the corresponding corrupted input. The denoiser $f_\theta$ receives the noised
sequence together with the per-token times and predicts the clean target $\hat{x}_1$.

A single scalar $p \equiv p_\text{condclean}$ controls the probability of corruption of conditioning tokens.
For each example we draw a shared time $t \sim \mathcal{U}(0,1)$ and either have $\emph{all}$ the 
conditioning tokens match the noise level of the response tokens or have $\emph{all}$ remain clean: 
\[
\begin{gathered}
c \sim \mathrm{Bernoulli}(p), \qquad \tau_i = t \;\; (i \in \mathcal{G}), \\[4pt]
\tau_i =
\begin{cases}
1 & \text{if } c = 1 \;\; (\text{clue clean}),\\
t & \text{if } c = 0 \;\; (\text{clue noised}),
\end{cases}
\quad (i \in \mathcal{C}).
\end{gathered}
\]
The training objective is the cross-entropy between the prediction and the original sequence over the
loss region $\mathcal{L}$ (the full sequence):
\[
\mathcal{L}(\theta) \;=\; \mathbb{E}_{x_1,\, t,\, \{\tau_i\}}
\Big[\, \textstyle\sum_{i \in \mathcal{L}} \mathrm{CE}\big(f_\theta(x(\tau))_i,\; (x_1)_i\big) \Big].
\]
Setting $p=1$ recovers the standard ``paste-in'' conditional objective (conditioning tokens always clean);
$p=0$ recovers a fully unconditional joint distribution objective.\footnote{We note that at $p=0$, at intermediate
noise levels, the prompt therefore retains some signal.}
Intermediate $p$ interpolates between the two objectives, allowing the model to train under 
a range of conditioning strengths while also preventing a large training/inference gap. 

\subsection{Sampling}
\label{sec:method-sampling}
We present results under two inference strategies. First, we present our main results 
by following standard practice for conditional flow language models: we keep the conditioning
input clean and clamped to their clean one-hot values and moreover re-pasted into the 
sequence after every solver step. 
We integrate with a first-order Euler sampler. Since sampling assumes conditioning tokens are kept
clean, a small fraction $p$ of training examples with clean conditioning tokens is necessary 
to maintain generation performance, as demonstrated in Section~\ref{sec:ablations}.   

Second, in Section~\ref{sec:cfg-sampling} we present results using guided 
sampling. Since our noised-prompt objective trains our model with both clean and noised prompts, 
we take inspiration from classifier-free guidance~\citep{cfg} and sample using an affine combination
of the model's prediction with a noised and clean prompt. Specifically we combine the log-softmax
logits of the model's $\hat{x}_1$ simplex prediction with a guidance weight $w$ to get an 
affine interpolation $l = f_u + w(f_c - f_u)$—where 
$f_u = f(x_t, y_t)$ and $f_c = f(x_1, y_t)$ represent the model's predictions under 
noised and clean prompts respectively—and softmax our combined logits to get a combined 
$\hat{x}_1$ simplex prediction for sampling. By choosing $w$ we control
how much the model relies on the prompt during inference.

\section{Experiments}
\label{sec:experiments}
\begin{figure}[t]
  \centering
  \includegraphics[width=0.95\linewidth]
  {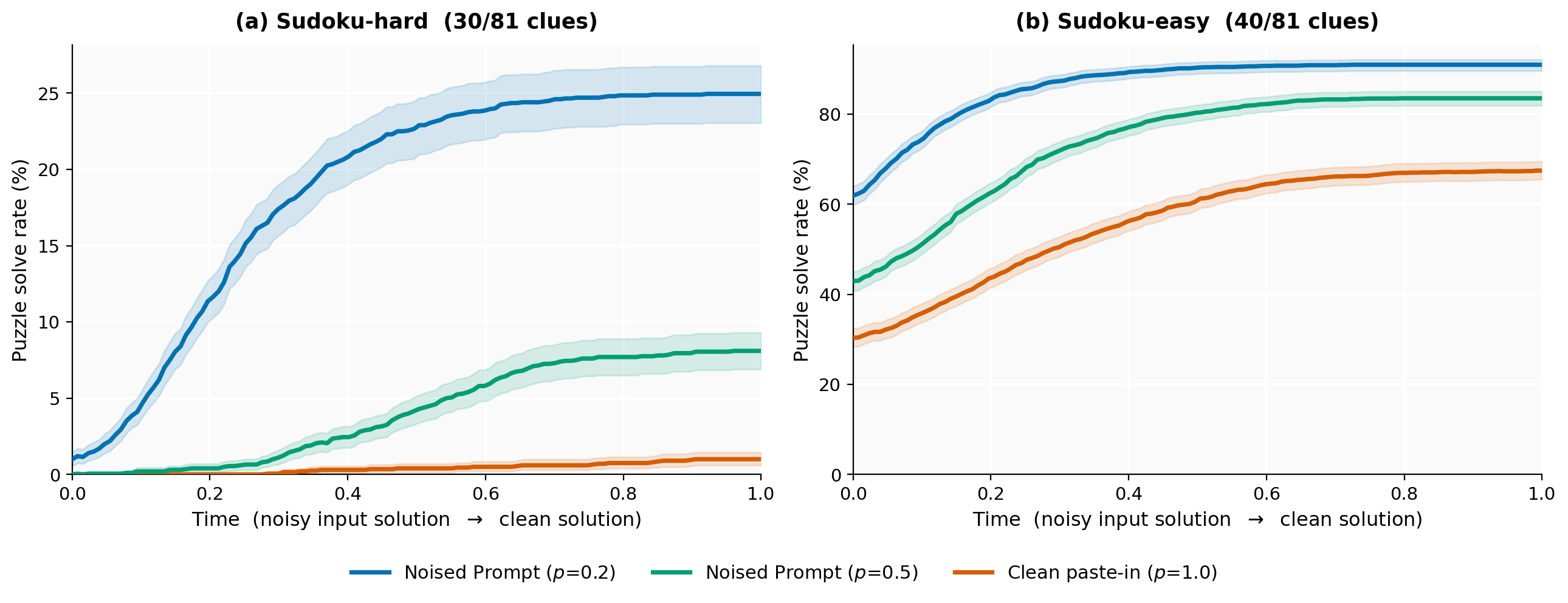}
\caption{
\textbf{Puzzle solve rate over the sampling trajectory.}
Puzzle (exact match) solve rate as a function of sampling time
(noisy input solution $\rightarrow$ clean solution), read from the model's $x_1$
prediction at each step. Means calculated over $2000$ held-out puzzles with a $128$-step
Euler sampler; shaded bands are single-seed $95\%$ bootstrap intervals over puzzles.
Across both difficulties the noised-prompt models ($p_\text{condclean}{=}0.2, 0.5$)
reach a substantially higher solve rate than the standard paste-in model
($p_\text{condclean}{=}1.0$) and also saturate their solve rate bound earlier in generation.
}
\label{fig:solverate}
  \end{figure}

\subsection{Setup}
\label{sec:setup}
Across all experiments, we use the Flow Language Model (FLM) from \cite{flm}. 
Unless otherwise specified, the backbone is a DiT-style transformer \citep{peebles2023dit} with hidden size $512$,
$8$ blocks, $8$ attention heads, conditioning dimension $128$, and dropout $0.1$, trained with a
flow objective and $x_1$-prediction. We optimize with
AdamW (learning rate $3\times10^{-4}$, $(\beta_1,\beta_2)=(0.9,0.999)$, no weight
decay), a constant schedule with $2{,}500$ warmup steps, gradient clipping at $1.0$,
bf16-mixed precision, and an EMA of $0.9999$. We follow FLM and use antithetic time sampling and a $t$-reparametrization
technique that adjusts for the vocabulary size \citep{flm}. 
We sample with a $128$-step Euler solver.
For each experiment, the single swept variable is
$p_\text{condclean}$; all other settings are held
fixed. Full hyperparameters are in Appendix~\ref{app:hyperparams}. All runs train on a single
NVIDIA L40S GPU.

\paragraph{Combinatorial reasoning tasks.}
For Sudoku, the vocabulary designates token $0$ as
PAD, token $1$ an empty cell, and tokens $2$--$10$ as the digits $1$--$9$; for N-Queens,
token $0$ is PAD, token $1$ is an empty cell, and token $2$ is a queen.
Following prior work \citep{hyperspherical_flows}, we train with no data augmentation.

\paragraph{Natural language tasks.}
Both language tasks (Section~\ref{sec:language}) use the same bert-base-uncased tokenizer
with ($V{=}30{,}522$). They also use the same format (\texttt{[CLS] source [SEP] target [EOS]}), padded to a fixed length.
Each evaluation is on a held-out set of $1{,}000$ prompts.

\paragraph{Evaluation checkpoint.}
On Sudoku and N-Queens we on evaluate using the final checkpoint for a matched compute comparison
and because the models do not overfit within their training budget. On the other hand, 
the language models overfit early (Section ~\ref{sec:valnll}) and therefore we evaluate 
using the best validation NLL checkpoint, with validation loss validated every $2{,}000$ steps.   

\subsection{Metrics}
\label{sec:method-metrics}
\paragraph{Combinatorial reasoning.}
For Sudoku and N-Queens we report two metrics. \textbf{Accuracy} is the fraction of generated boards that are valid
solutions to the puzzle (aggregated across all puzzles and samples). \textbf{Coverage} (N-Queens) is
a per puzzle statistic capturing the fraction of \emph{distinct} valid solutions
recovered across $k=20$ generated samples, all conditioned on the same input puzzle. This
allows us to measure the diversity of the successful generations.

\paragraph{Natural language.}
We report a diversity-quality frontier, as quality metrics can be gamed at the cost of diversity (\cite{pynadath2026generativefrontiersevaluationmatters},
\cite{replaid}, \cite{elf}).
We empirically find that generative perplexity as scored by GPT-2 (\cite{gpt2}) rewards degenerate repeating sequences and so use
\textbf{MAUVE} \citep{mauve}, which scores how well the distribution of generated text 
matches the human reference text. We compute MAUVE of the generated responses against the human data responses using $1{,}000$ samples.    
\textbf{Average unigram entropy} measures the diversity of the generated text; we add the reference 
entropy values ($2.935$ bits on Gigaword and $3.878$ on dialogue) as a reference target.

\paragraph{Confidence Intervals.}
Unless otherwise stated, we report $95\%$ two-sided $t$-distribution confidence intervals across $n=5$ training seeds
(dialogue uses $n=3$ training seeds). We note that all training seeds are scored on the same evaluation set
 and only vary in initialization and batch order. We clip the CI to be non-negative.
Bootstrap confidence intervals ($95\%$) are obtained by resampling the
evaluation puzzles ($10{,}000$ resamples for Sudoku).
\subsection{Sudoku}
\label{sec:sudoku}
Sudoku is a constraint-satisfaction puzzle on a $9\times9$ grid and a common benchmark for
reasoning and generalization ~\citep{trm, gram}. We use it precisely to measure generalization:
to solve unseen puzzles a model cannot simply memorize a mapping from $x \to y$, 
it must implicitly learn and respect the constraints of the game.

We train on Sudoku with the in-place infill layout: a full $9\times9$ board flattened row-wise (length $81$), with the
loss taken over all board tokens. Puzzles are generated following ~\citet{hyperspherical_flows}, and
difficulty is controlled by the number of revealed clues---$40$ of $81$ cells given for
\emph{easy} and $30$ of $81$ for \emph{hard}. We sweep $p_\text{condclean}$ at two training-set
sizes, $N=10{,}000$ and $N=1{,}000$ puzzles and evaluate accuracy on $2{,}000$ held-out puzzles with a $128$-step sampler
(Table~\ref{tab:sudoku}).

\begin{table}[t]
  \centering
  \caption{Sudoku solve rate (\%) on $2{,}000$ held-out puzzles under the conditioning clean
  probability $p_\text{condclean}$.}
  \label{tab:sudoku}
  \begin{threeparttable}
    \begin{tabular}{lcccc}
    \toprule
    & \multicolumn{2}{c}{Easy ($40/81$ clues)} & \multicolumn{2}{c}{Hard ($30/81$ clues)} \\
    \cmidrule(lr){2-3}\cmidrule(lr){4-5}
    $p_\text{condclean}$ & $N=10{,}000$ & $N=1{,}000$ & $N=10{,}000$ & $N=1{,}000$ \\
    \midrule
    $0.0$ (uncond.) & $1.41$ & $0.00$ & $0.00$ & $0.00$ \\
      & {\scriptsize $[0.03,\,2.79]$}& {\scriptsize $[0.00,\,0.00]$} & {\scriptsize $[0.00,\,0.00]$} & {\scriptsize $[0.00,\,0.00]$} \\
    $0.1$ & $88.59$ & $2.42$ & $\mathbf{24.65}$ & $0.02$ \\
      & {\scriptsize $[85.88,\,91.30]$} & {\scriptsize $[0.00,\,4.99]$} & {\scriptsize $[18.56,\,30.74]$} & {\scriptsize $[0.00,\,0.08]$} \\
    $0.2$ & $88.89$ & $1.39$ & $24.25$ & $0.01$ \\
      & {\scriptsize $[84.81,\,92.97]$} & {\scriptsize $[0.11,\,2.67]$} & {\scriptsize $[20.04,\,28.46]$} & {\scriptsize $[0.00,\,0.04]$} \\
    $0.3$ & $\mathbf{91.18}$ & $0.71$ & $22.55$ & $0.07$ \\
      & {\scriptsize $[86.72,\,95.64]$} & {\scriptsize $[0.00,\,1.92]$} & {\scriptsize $[17.72,\,27.38]$} & {\scriptsize $[0.00,\,0.19]$} \\
    $0.5$ & $87.51$ & $\mathbf{6.92}$ & $14.70$ & $\mathbf{0.21}$ \\
      & {\scriptsize $[80.93,\,94.09]$} & {\scriptsize $[0.86,\,12.98]$} & {\scriptsize $[8.19,\,21.21]$} & {\scriptsize $[0.16,\,0.26]$} \\
    $1.0$ (paste-in) & $68.49$ & $1.47$ & $3.73$ & $0.00$ \\
      & {\scriptsize $[57.44,\,79.54]$} & {\scriptsize $[0.00,\,3.11]$} & {\scriptsize $[2.76,\,4.70]$} & {\scriptsize $[0.00,\,0.00]$} \\
    \bottomrule
    \end{tabular}
    \end{threeparttable}
  \end{table}

The default setup of pasting-in clean conditioning tokens during training performs 
notably worse than our method of noising conditioning tokens. 
On Sudoku easy, $p{=}0.3$ reaches $91.18\%$ versus
$68.49\%$ for the standard paste-in setup; on hard with $p{=}0.1$, $24.65\%$ versus $3.73\%$. 
Under the data-limited regime $N=1{,}000$, $p{=}0.5$ performs best; 
the default paste-in strategy and $p{=}0.2$ perform similarly, with no statistically significant
difference in performance. 

The fully-unconditional setup ($p=0.0$) exhibits poor generation, demonstrating that the model needs to see
some clean prompts so our inference time clean prompt is within the training distribution.

\subsection{N-Queens}
\label{sec:nqueens}
N-Queens places $n$ mutually non-attacking queens on an $n \times n$ board and likewise evaluates reasoning and generalization~\citep{gram}. 
In addition to being combinatorial in nature,
N-Queens admits multiple valid solutions for a given starting board. This allows us to measure the
diversity of the generated solutions. Following GRAM~\citep{gram}, we track coverage of possible
solutions and report the fraction of unique solutions generated out of the total number of possible solutions corresponding to
that board. We evaluate on held out initial board states and generate 20 samples per input
board. 

We consider two scales, $n=8$ (length $64$) and $n=10$ (length $100$). We subsample the data
such that puzzles of varying difficulty (as defined by the number of initial queens) are equal
in number. For $n=8$ this yields $1{,}758$ training / $174$ validation samples; for
$n=10$ we have $15{,}000$ training / $1{,}671$ validation samples. We also hold-out a seperate
evaluation partition of $750$ input boards for both $n=8$ and $n=10$. Training and evaluation splits are
disjoint by input board. We train at global batch size $256$ for $100$k
steps and evaluate the final checkpoint on $750$ held-out puzzles with $20$ generations each,
repeating every configuration over $5$ training seeds (Table~\ref{tab:nqueens}).

\begin{table}[t]
\centering
\caption{N-Queens infill accuracy and coverage (\%), final checkpoint, $750$ puzzles
$\times\,20$ samples.}
\label{tab:nqueens}
\begin{tabular}{lcccc}
\toprule
& \multicolumn{2}{c}{$n=8$ (length 64)} & \multicolumn{2}{c}{$n=10$ (length 100)} \\
\cmidrule(lr){2-3}\cmidrule(lr){4-5}
$p_\text{condclean}$ & Accuracy & Coverage & Accuracy & Coverage \\
\midrule
$0.2$ & $\mathbf{99.37}$ & $\mathbf{92.39}$ & $\mathbf{98.32}$ & $\mathbf{73.79}$ \\
 & {\scriptsize $[99.07,\,99.66]$} & {\scriptsize $[91.92,\,92.86]$} & {\scriptsize $[97.91,\,98.73]$} & {\scriptsize $[73.19,\,74.39]$} \\
$0.5$ & $98.67$ & $90.48$ & $95.09$ & $61.12$ \\
 & {\scriptsize $[98.27,\,99.06]$} & {\scriptsize $[89.59,\,91.38]$} & {\scriptsize $[94.55,\,95.63]$} & {\scriptsize $[60.59,\,61.64]$} \\
$1.0$ (paste-in) & $97.48$ & $89.02$ & $86.23$ & $50.60$ \\
 & {\scriptsize $[96.73,\,98.23]$} & {\scriptsize $[88.45,\,89.59]$} & {\scriptsize $[83.67,\,88.78]$} & {\scriptsize $[49.41,\,51.80]$} \\
\bottomrule
\end{tabular}
\end{table}

At $n=8$ accuracy is high throughout all $p$; whereas at $n=10$ noising the conditioning
tokens ($p=0.2$) performs substantially better than the default paste-in setup
($98.32$ vs $86.23\%$). 

Coverage is a more salient metric: for both $n=8$ and $n=10$, training on just clean conditioning
tokens leads to lower diversity of generations, prominently demonstrated at $n=10$ ($50.60$ vs
$73.79\%$). We report additional diversity results in Appendix \ref*{app:diversity}.

\subsection{Natural Language Generation}
\label{sec:language}
We also present results on conditional language generation and specifically we consider the tasks of summarization 
and open ended dialogue. While we see benefits in summary generation—even when compared to regularized baselines—as
 measured by the mauve-diversity frontier, we do not find supporting evidence in open ended dialogue generation. 
 For natural language generation, we report results with categorical sampling of $\hat{x}_1$ distribution for all steps,
 whereas previously for Sudoku/N-Queens we argmax decode the last sampling step. 

\subsubsection{Gigaword summarization}
\label{sec:gigaword}
For summarization we use the Gigaword dataset \citep{gigaword}, which contains pairs of source text and
its summary headline. We filter to sequences of at most $L{=}80$ tokens whose headline is
at least three words, leaving $191{,}082$ training and $3{,}950$ validation pairs. We do an initial single 
seed sweep across $p_\text{condclean} \in \{0.3, 0.5, 0.8, 1.0\}$; every averaged seed result compares 
 $p{=}0.5$ against the paste-in baseline over $n{=}5$ training seeds. To compare noising prompts to standard 
 regularization methods, we additionally sweep over dropout probability, as reported in Figure~\ref{sec:valnll}.    

\subsubsection{Open dialogue}
\label{sec:dialogue}
Open dialogue benchmarks the natural language setting where the response is not as easily determined
from the prompt. We use the Commonsense Conversation Dataset \citep{ccd} as released and processed by 
the DiffuSeq codebase \citep{diffuseq}; sequences are padded to $L{=}96$. We train both on the original 
$3.38$M training dataset and also a training subset of $191$k pairs (matching Gigaword's training set size), both with the DiffSeq's original $2{,}048$-pair validation split.
We sweep $p_\text{condclean} \in \{0.3, 0.5, 1.0\}$ over $3$ training seeds.

\subsubsection{Validation likelihood and overfitting}
\label{sec:valnll}
During training, for the validation set we always keep the prompt clean, so our 
validation likelihood is comparable across different $p_\text{condclean}$ probabilities. 
We observe that noising the prompt acts as a strong regularizer for smaller language datasets ($191$k samples),
providing robustness to overfitting (Table~\ref{tab:valnll-gigaword}); however we do not see a significant
regularization advantage for larger datasets (Table~\ref{tab:valnll-dialogue}). On Gigaword we see that
noising prompts yields better val NLL even compared to heavily regularized baseline (dropout $0.4$),
demonstrating its effectiveness over ordinary regularization. We demonstrate val NLL improvements on larger
(132M) models as well. 

\begin{table}[t]
  \centering
  \caption{Gigaword validation NLL. (mean $\pm$ std.\ over $5$ training seeds)}
  \label{tab:valnll-gigaword}
  \begin{tabular}{llccc}
  \toprule
  Model & $p_\text{condclean}$ & Dropout & Best Val NLL & Checkpoint Step \\
  \midrule
  $57$M  & $1.0$ (paste-in) & $0.1$ & $0.4581 {\scriptstyle\ \pm\ 0.0041}$ & $8{,}000$ \\
  $57$M  & $1.0$ & $0.2$ & $0.4537 {\scriptstyle\ \pm\ 0.0104}$ & $12{,}000$ \\
  $57$M  & $1.0$ & $0.3$ & $0.4478 {\scriptstyle\ \pm\ 0.0056}$ & $16{,}400$ \\
  $57$M  & $1.0$ & $0.4$ & $0.4455 {\scriptstyle\ \pm\ 0.0080}$ & $19{,}200$ \\
  $57$M  & $\mathbf{0.5}$ & $0.1$ & $\mathbf{0.3707} {\scriptstyle\ \pm\ 0.0093}$ & $23{,}600$ \\
  \midrule
  $132$M & $1.0$ (paste-in) & $0.1$ & $0.4573 {\scriptstyle\ \pm\ 0.0022}$ & $6{,}000$ \\
  $132$M & $\mathbf{0.5}$ & $0.1$ & $\mathbf{0.3518} {\scriptstyle\ \pm\ 0.0112}$ & $14{,}400$ \\
  \bottomrule
  \end{tabular}
\end{table}

\subsubsection{Diversity--quality frontier results}
\label{sec:frontier}
\begin{figure}[t]
  \centering
  \includegraphics[width=0.49\linewidth]{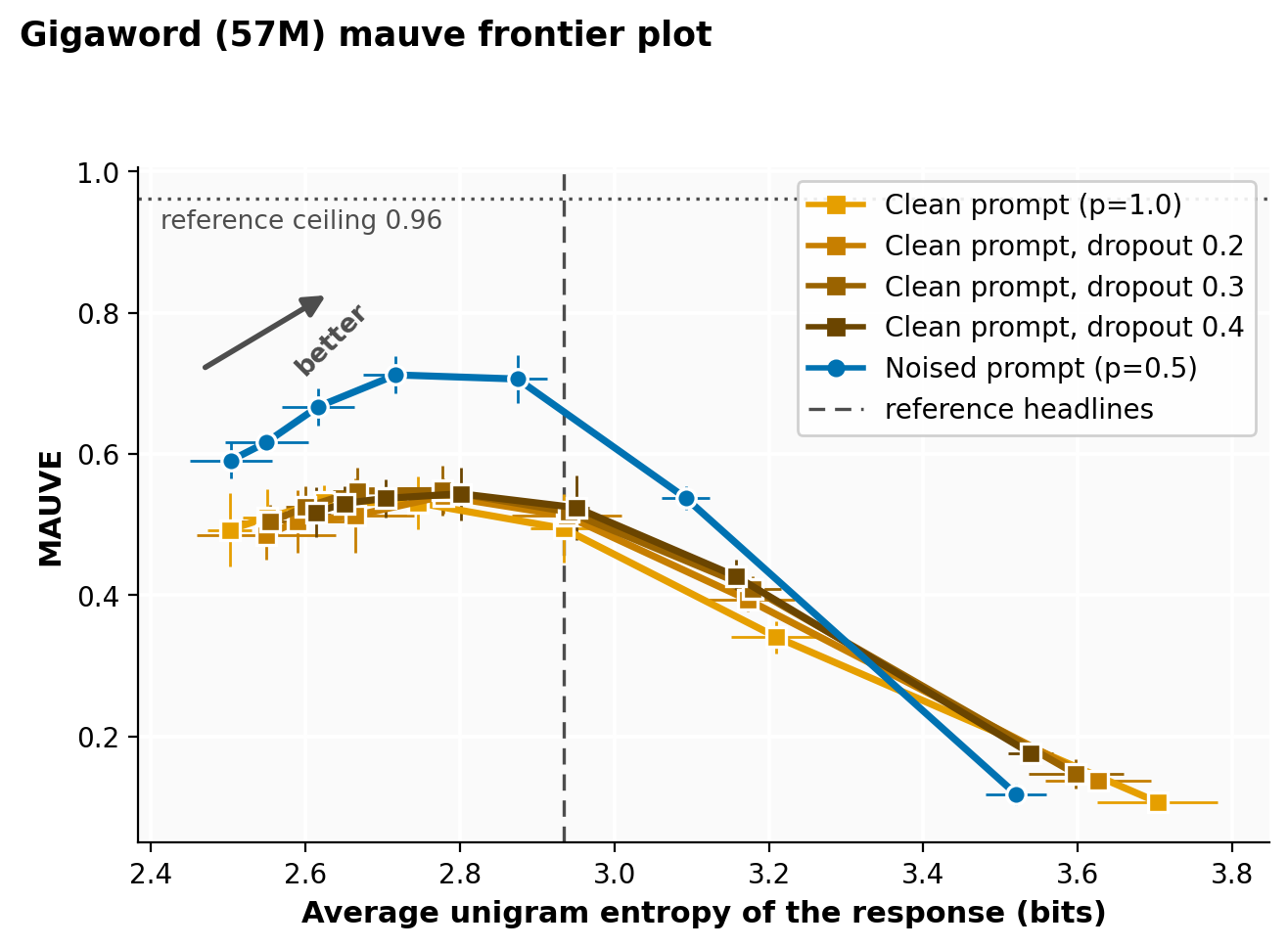}
  \includegraphics[width=0.49\linewidth]{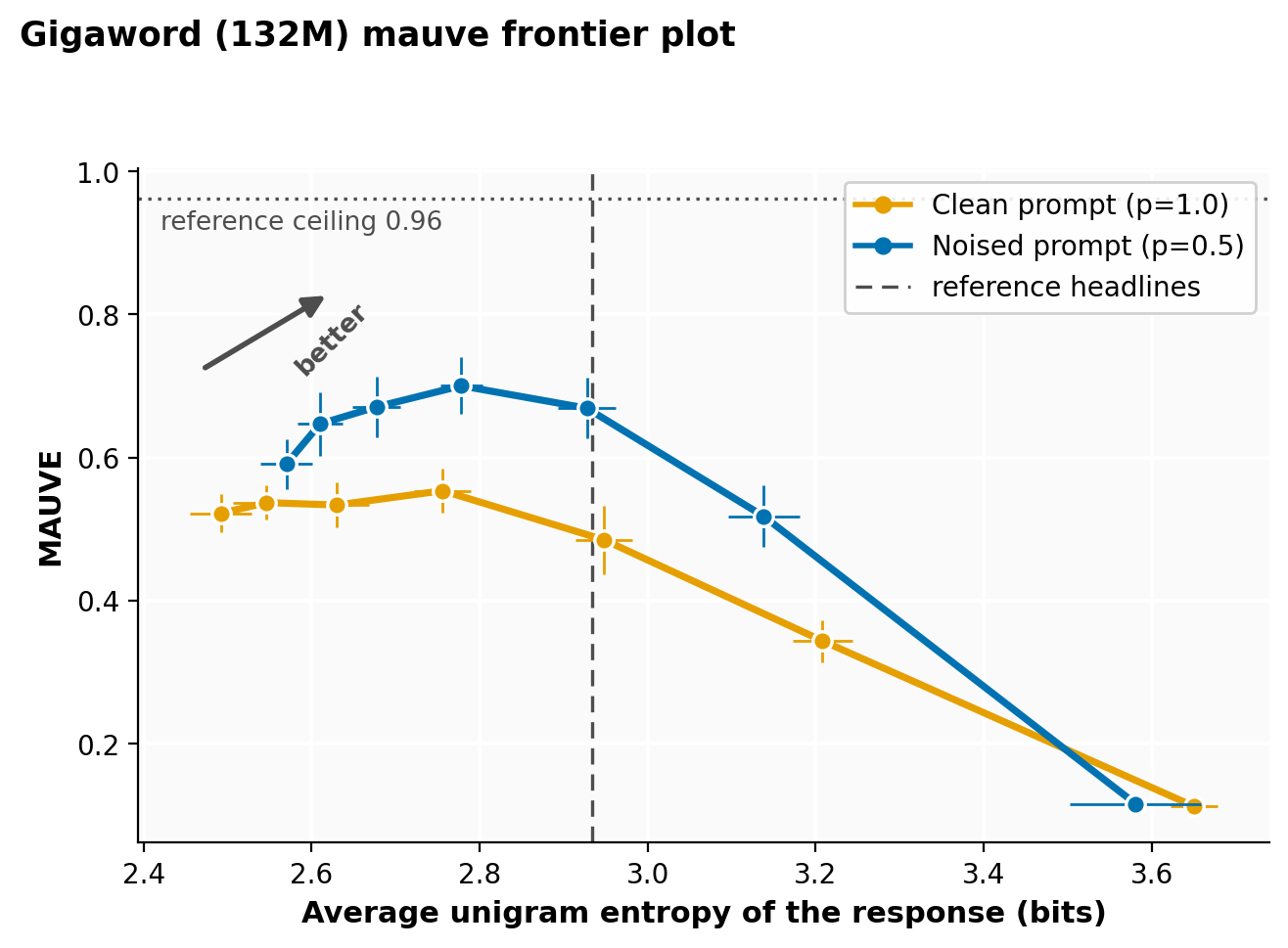}
  \includegraphics[width=0.49\linewidth]{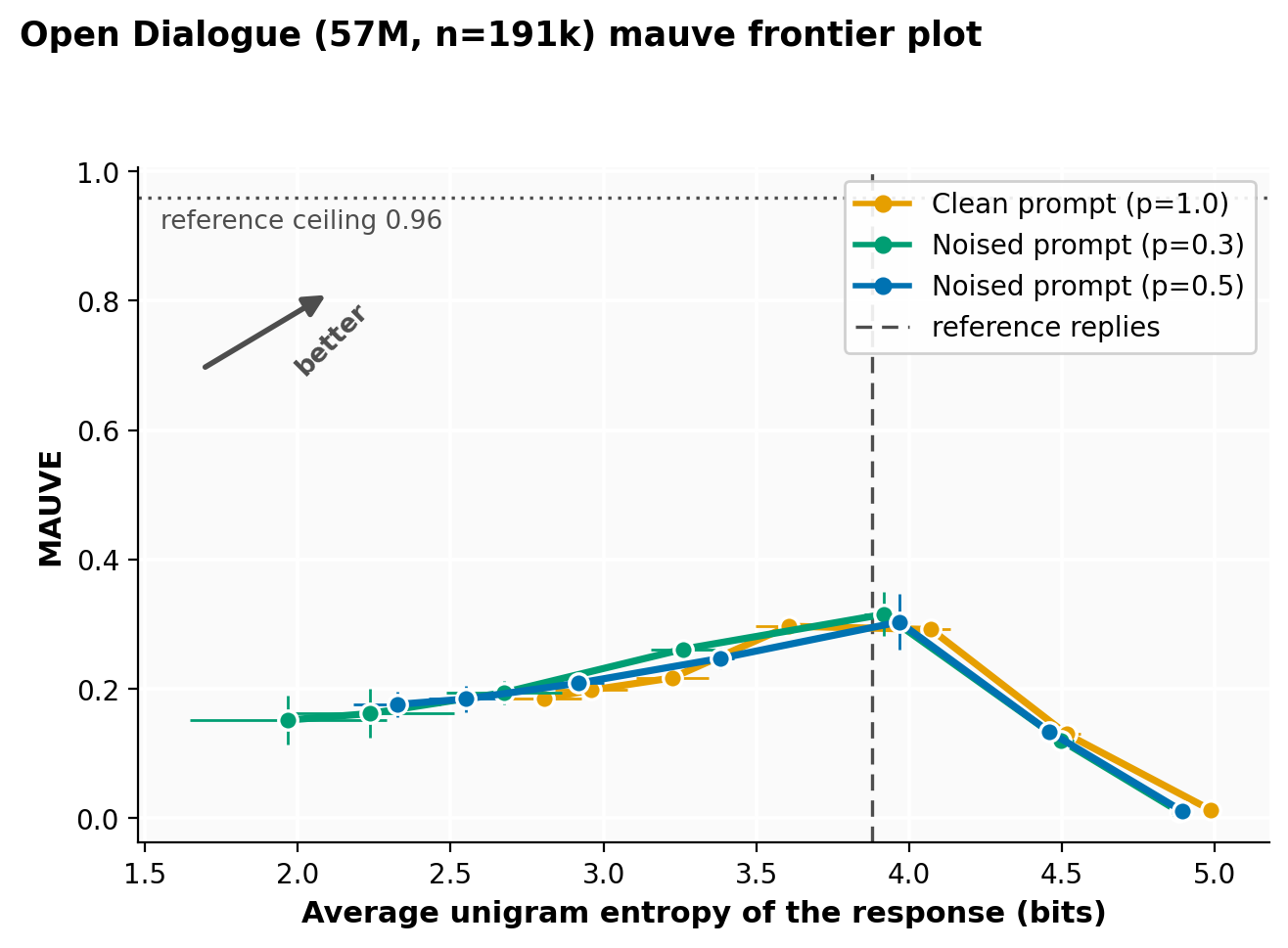}
  \includegraphics[width=0.49\linewidth]{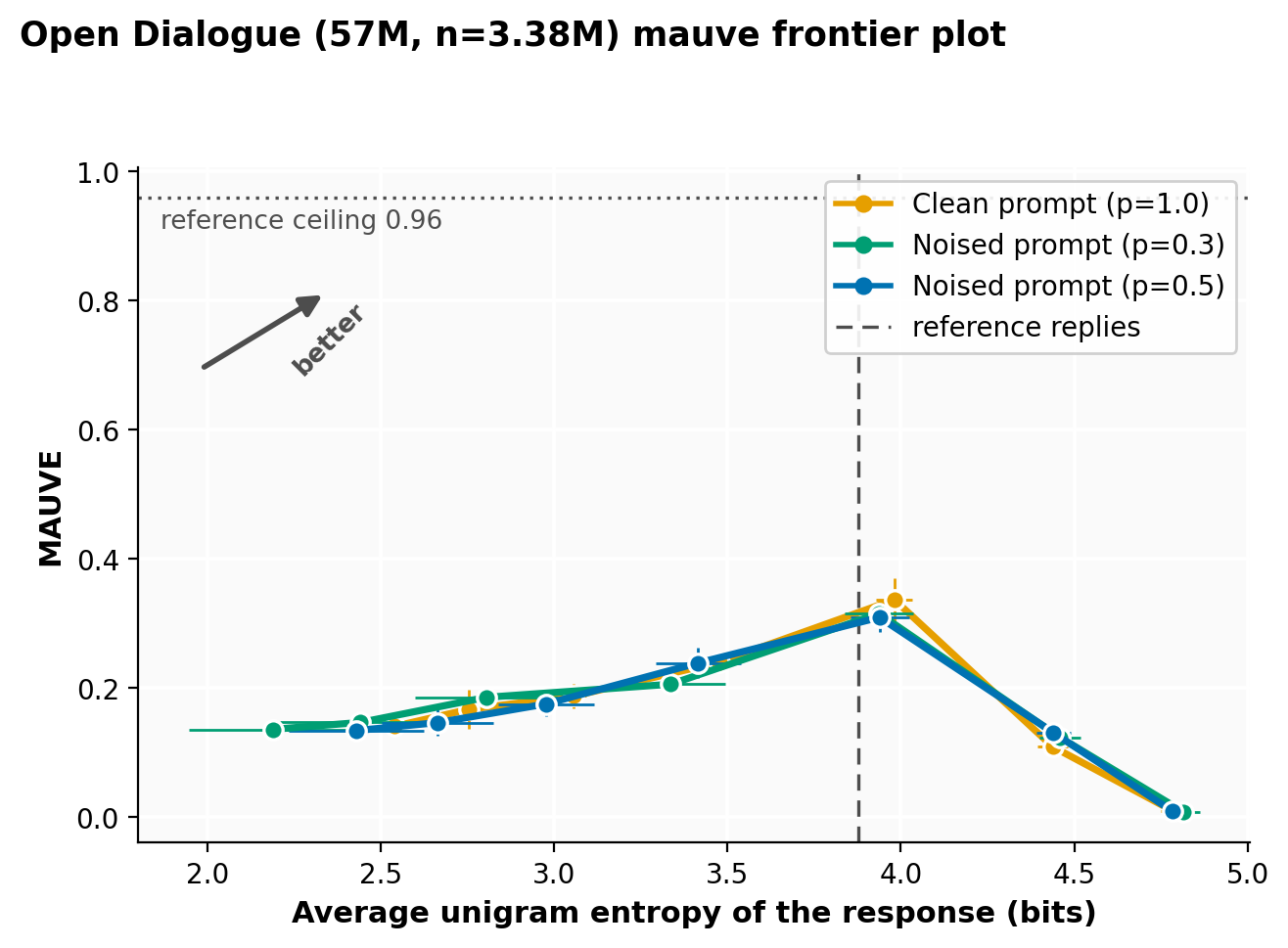}
  \caption{
  \textbf{Diversity--quality frontiers.} MAUVE against average response unigram entropy,
  traced by a softmax temperature on the denoised $x_1$ distribution ($T \in [0.2, 1.5]$).
  We provide as a target the human references MAUVE (estimated by scoring data halves against each other)
  and average unigram entropy. We note that the default dropout is $0.1$ and bars represent training
  seed standard deviation. 
  }
  \label{fig:frontier}
\end{figure}

Following commentary by \cite{pynadath2026generativefrontiersevaluationmatters} that diversity--quality frontier
plots are more accurate and faithful depictions of model generative quality, we report a diversity--quality frontier plot by sweeping over the temperature on
the denoised $x_1$ probability distribution (Figure~\ref{fig:frontier}). 
On Gigaword, noised prompt training dominates clean prompt training at and below reference target entropy, even with 
significant dropout regularization, demonstrating that noising prompts adds a benefit 
beyond standard regularization techniques. We also see that the improvement 
holds for a larger 132M model. 

On open dialogue, the story is different. We do not see a significant effect 
under noising, even at larger noising probabilities levels ($p{=}0.3$). We see no significant effect 
for both the larger data (3.38M samples) regime and smaller data (191k samples) regime. 

\subsection{Guided sampling}
\label{sec:cfg-sampling}
Because our objective trains the model under both clean and noised conditioning tokens, we can
sample with guided sampling described in Section~\ref{sec:method-sampling}. We affine interpolate
the model's predictions with noised and clean prompts, $l = f(x_t, y_t) + w(f(x_1, y_t) - f(x_t, y_t))$ before applying a softmax
to get our guided $\tilde{x}_1$ simplex prediction to use during inference. We note that $w{=}1$ recovers
the default paste-in sampling procedure, and larger $w > 1$ encourages the model to rely more on the 
prompt information during inference. 
We add full details in Appendix~\ref{app:cfg-sweep}.

\begin{table}[t]
  \centering
  \small
  \setlength{\tabcolsep}{4pt}
  \caption{Sudoku solve rate (\%) under guided sampling at $p_\text{condclean}{=}0.2$ on 2,000
  held-out puzzles. $w{=}1$ represents default sampling with a clean prompt during inference time.}
  \label{tab:cfg-sudoku}
  \begin{tabular}{llcccc}
  \toprule
  & & \multicolumn{2}{c}{Easy} & \multicolumn{2}{c}{Hard} \\
  \cmidrule(lr){3-4}\cmidrule(lr){5-6}
  Setting & $w$ & Mean & $\Delta$ (improvement) & Mean & $\Delta$ (improvement) \\
  \midrule
  \multirow{4}{*}{$N{=}10{,}000$}
    & $1$ & $89.44$ & --- & $23.89$ & --- \\
    & & {\scriptsize $[85.62,\,93.26]$} & & {\scriptsize $[20.25,\,27.53]$} & \\
    & $5$ & $\mathbf{92.08}$ & $+2.64$ & $\mathbf{30.82}$ & $+6.93$ \\
    & & {\scriptsize $[88.79,\,95.37]$} & {\scriptsize $[1.83,\,3.45]$} & {\scriptsize $[26.86,\,34.78]$} & {\scriptsize $[6.55,\,7.31]$} \\
  \midrule
  \multirow{4}{*}{$N{=}1{,}000$}
    & $1$ & $1.66$ & --- & $0.03$ & --- \\
    & & {\scriptsize $[0.09,\,3.23]$} & & {\scriptsize $[0.00,\,0.09]$} & \\
    & $3$ & $\mathbf{3.94}$ & $+2.28$ & $0.03$ & $+0.00$ \\
    & & {\scriptsize $[0.24,\,7.64]$} & {\scriptsize $[0.14,\,4.42]$} & {\scriptsize $[0.00,\,0.11]$} & {\scriptsize $[0.00,\,0.12]$} \\
  \bottomrule
  \end{tabular}
\end{table}

On Sudoku, which has a unique solution per puzzle, guidance offers an improvement to performance,
increasing solve rate accuracy, with the gains most visible on harder tasks.\footnote{The $w{=}1$ rows use the same clean prompt sampling
procedure and the same $p{=}0.2$ checkpoints as in Tables~\ref{tab:sudoku}.
Sampling noise makes the re-evaluated numbers differ slightly from those reported earlier.} On Sudoku-hard ($N{=}10{,}000$) 
guidance increases accuracy by $6.93$ ($23.89 \to 30.82$) and in the data-limited $N{=}1{,}000$ easy
setting it more than doubles the solve rate ($1.66 \to 3.94$). 
However, as reported in ~\citep{cfg}, guidance presents a trade off with solution diversity, which is 
apparent with N-Queens (Table~\ref{tab:cfg-nqueens-seeds}). While increasing guidance generally improves accuracy, it 
decreases solution diversity, as measured by coverage (Table~\ref{tab:cfg-nqueens}).

\subsection{Ablations}
\label{sec:ablations}
We consider the following ablations to illustrate the design choice of our noising scheme:
(i) whether the conditioning tokens are noised to the same time as the response or to an independent time;
(ii) whether the loss is computed over the full sequence or only the response tokens; (iii) the
whether the model needs to see clean conditioning tokens at all (fully unconditional training); (iv) the 
choice of gaussian noise for prompt tokens versus masked token noise; and (v) noising the prompt jointly. 

\begin{figure}[!tbp]
  \centering
  \includegraphics[width=0.6\linewidth]
  {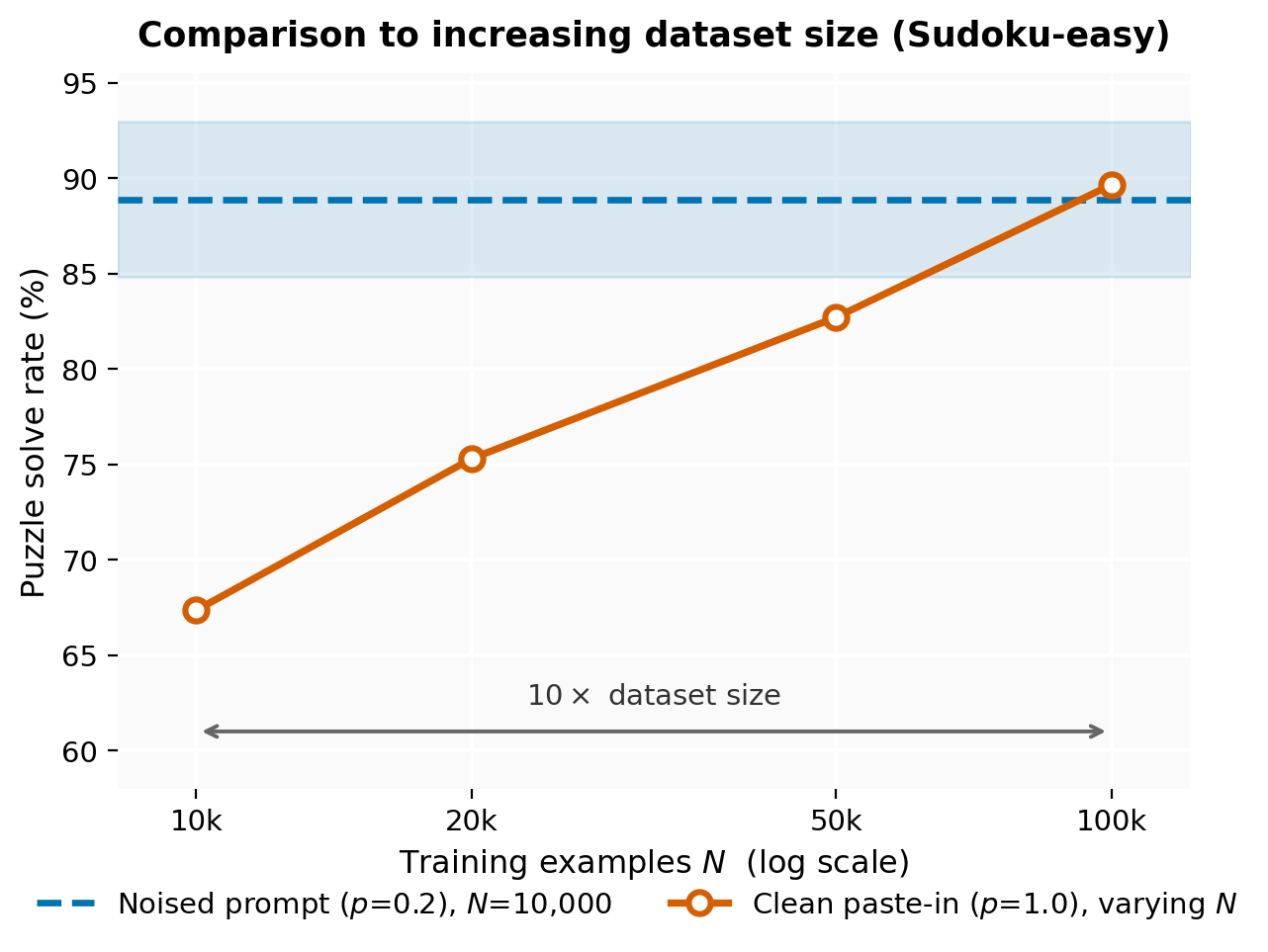}
\caption{
\textbf{Comparison to scaling dataset size.}
Puzzle (exact match) solve rate on Sudoku-easy using default paste-in method trained 
under increasing dataset sizes. All training datasets are subsets of a $100$k training dataset.
The default paste-in training objective requires $10\times$ more data to match the performance 
under noised prompt training $p{=}0.2$.   
}
\label{fig:datascaling}
  \end{figure}

Moreover, to demonstrate the efficacy our method, we compare performance gains 
relative to increasing the number of training samples. All numbers are calculated with $95\%$ two-sided t-distribution CIs using $5$ training seeds,
excluding the data scaling numbers in Figure~\ref{fig:datascaling}, which is calculated using a single seed. 

\paragraph{Separate conditioning time.}
The current objective noises conditioning tokens to the same sampled noise level (parameterized by time) as the response tokens.
We consider drawing an independent time $t' \sim \mathcal{U}(0,1)$ for the conditioning tokens, so
the conditioning and response are corrupted to decoupled noise levels. 

On Sudoku easy ($N{=}10{,}000$) it reaches
$93.23\%$, above the best single-time setting in Table~\ref{tab:sudoku}. ($91.18\%$ at $p{=}0.3$),
while on Sudoku-hard it performs similarly at $25.09\%$. On N-Queens it performs worse than
frequent same time noising (Table~\ref{tab:nqueens}).

\paragraph{Loss calculated on full sequence vs response only.}
We ablate a response-tokens only loss objective that excludes the conditioning tokens predictions,
sweeping $p_\text{condclean}$ on Sudoku-easy ($N{=}10{,}000$). By incorporating only the response 
tokens in the loss, the model performs similar or worse, offering support for our position 
that a full sequence objective is a more effective training objective. 

Notably, the noised prompt training still significantly
outperforms paste-in clean prompt training ($87.70\%$ at $p{=}0.2$ vs $63.34\%$ at $p{=}1.0$),
demonstrating that the performance gains mainly come from noising, rather than a full sequence
loss objective.

\paragraph{Only noised conditioning tokens ($p_\text{condclean}{=}0$).}
With conditioning tokens always noised, the model learns the unconditional joint distribution 
but is unable to utilize the given clean conditioning information at inference time. On Sudoku hard
 ($N{=}10{,}000$) accuracy collapses to $0\%$; and on Sudoku easy ($N{=}10{,}000$) accuracy is $1.41\%$.
This confirms that some exposure to clean conditioning tokens is necessary during training.

\paragraph{Masked token noise.}
Currently our method noises the prompt token using gaussian noise, the default  
choice of continuous diffusion language models. We also evaluate whether our method can hold
under other types of noise, such as masked token noise. As before for some probability $p_\text{condclean}$ the sequence has a fully clean prompt; 
else, we noise the prompt sequence, but this time we replace a matched number of the prompt tokens to 
[MASK], removing all signal from those tokens. We choose the fraction of masked tokens to match the amount of
information lost via gaussian noising at level $t$.

Specifically, assuming a uniform prior over token IDs with a vocabulary $V$, 
we can characterize the mutual information of a one-hot noised token $y$ with its token ID $c$. 
$$MI(t) = \log{V} - H(c|y)$$ Then we can define the ratio of mutual information 
to the max mutual information, $r(t) = \frac{MI(t)}{\log{V}}$, and we can set our masking probability 
$m(t) = 1 - r(t)$ such that the ratio of mutual information under masking matches the ratio of 
mutual information under gaussian noise. We see that for both $p{=}0.2$ and $p{=}0.5$ noising the prompt via masked noise performs statistically
similar to gaussian noising, meaning that our method can extend to other noising forms as well.

\paragraph{Individual noising.}
We ablate the decision to either keep the prompt clean or to noise the prompt jointly by instead 
considering per token bernoulli noising for all sequences, i.e each token remains clean based on an independent
draw with $p_\text{condclean}$. We note that this causes a train/inference gap with clean prompt sampling and 
correspondingly the method is able to perform better at the higher clean probability $p=0.5$ over $p=0.2$, albeit
overall worse than joint noising on Sudoku-hard. 

\paragraph{Comparison to data scaling.}
By also noising the prompt, our method can be seen as a form of data augmentation. 
Here we quantify the performance gains relative to increasing the number of training samples. 
We evaluate on Sudoku-easy and train a single seed on various size subsets of a $100$k training dataset, with a held-out 
set of $n=2,000$ samples drawn i.i.d from the same data generator used for previous Sudoku held-out sets.   
All training runs follow the same settings as in Appendix~\ref{app:hyperparams} and are computed matched. 
As seen in Figure~\ref{fig:datascaling}, training with a noised-prompt has the same effective performance as
increasing the total training dataset size by $10\times$ under the paste-in training objective, demonstrating 
the efficacy of our training method.

\begin{table}[t]
\centering
\caption{Ablations. \emph{First:} separate
conditioning time variant (independent clue time $t'$).\emph{Second:} Per token noising. \emph{Third:} response-only loss
region, swept over $p_\text{condclean}$. \emph{Fourth:}
Fully unconditional generation. \emph{Fifth:} Masked noise variant.} 
\label{tab:ablations}
\begin{tabular}{llcc}
\toprule
Ablation & Setting & Accuracy & Coverage \\
\midrule
\multirow{4}{*}{Separate cond.\ time}
 & Sudoku easy  & $93.23$ {\scriptsize $[91.95,94.51]$} & --- \\
 & Sudoku hard  & $25.09$ {\scriptsize $[21.66,28.52]$} & --- \\
 & N-Queens $n{=}8$ & $97.20$ {\scriptsize $[96.67,97.74]$} & $88.84$ {\scriptsize $[88.04,89.64]$} \\
 & N-Queens $n{=}10$ & $88.56$ {\scriptsize $[87.74,89.39]$} & $51.69$ {\scriptsize $[50.89,52.49]$} \\
\midrule
\multirow{6}{*}{Per-token Bernoulli}
 & Sudoku easy, $p{=}0.2$ & $75.97$ {\scriptsize $[72.89,79.05]$} & --- \\
 & Sudoku easy, $p{=}0.5$ & $89.75$ {\scriptsize $[85.46,94.04]$} & --- \\
 & Sudoku hard, $p{=}0.2$ & $8.99$ {\scriptsize $[6.49,11.49]$} & --- \\
 & Sudoku hard, $p{=}0.5$ & $16.98$ {\scriptsize $[12.88,21.08]$} & --- \\
\midrule
\multirow{4}{*}{\shortstack[l]{Response-only loss\\(Sudoku easy)}}
 & $p{=}0.0$ & $1.22$ {\scriptsize $[0.19,2.25]$} & --- \\
 & $p{=}0.2$ & $87.70$ {\scriptsize $[84.62,90.78]$} & --- \\
 & $p{=}0.5$ & $82.91$ {\scriptsize $[77.54,88.28]$} & --- \\
 & $p{=}1.0$ & $63.34$ {\scriptsize $[58.68,68.00]$} & --- \\
\midrule
\multirow{2}{*}{Unconditional ($p{=}0$)}
 & Sudoku easy & $1.41$ {\scriptsize $[0.03,2.79]$} & --- \\
 & Sudoku hard & $0.00$ {\scriptsize $[0.00,0.00]$} & --- \\
\midrule
\multirow{2}{*}{\shortstack[l]{Masking noise\\(Sudoku easy)}}
 & $p{=}0.2$ & $84.85$ {\scriptsize $[70.72,98.98]$} & --- \\
 & $p{=}0.5$ & $86.63$ {\scriptsize $[82.67,90.59]$} & --- \\
\bottomrule
\end{tabular}
\end{table}

\section{Analysis and Discussion}
\label{sec:analysis}
\paragraph{Why does noising the conditioning help?}
We posit views consistent with the results above. First, noising the conditioning tokens 
leads to a wider set of training inputs (seen as data augmentation) and also makes use of the conditioning token predictions 
in the loss. Second, by decreasing the amount of signal in conditioning tokens, the model becomes
less reliant on predicting responses mainly based on the condition, leading to a greater 
variety of generations, as exhibited by the mode coverage results (always-clean training ($p=1$) collapses toward fewer solutions).
However, we find that actually noising the conditioning tokens is critical (Appendix~\ref{sec:noise-time-ablation}).
Finally, incorporating both noisy conditional tokens and non-noisy conditioning tokens can be 
seen as expanding the learning objective: $p<1$ mixes an easier prediction task
with a harder prediction task. 

\paragraph{Limitations.}
We show significant improvements to reasoning on small combinatorial puzzle datasets with small flow language models.
We show mixed results on natural language generation, with improvements in modest summarization language datasets with 
small to medium sized flow language models, but null differences with open ended dialogue generation. Due to cost limitations, 
we have not evaluated on significantly larger models, but future work would include evaluating model and dataset scaling behavior. 

\FloatBarrier
\section{Conclusion}
\label{sec:conclusion}
We revisit an accepted default practice of keeping conditioning tokens clean during training of continuous 
diffusion language models and instead noise the conditioning tokens as well. We find noising
of conditioning tokens to be a simple yet effective technique with task-dependent benefits: it improves accuracy 
 and diversity on combinatorial reasoning puzzles and summarization generation quality on Gigaword,
  but does not improve generation quality on open-ended dialogue. We evaluate on relatively
   small models—future work includes scaling analysis with larger models and larger datasets.

\section*{Acknowledgements}
\label{sec:acknowledgements}
We thank Julian Quevado for his encouragement and support.
We thank the broader diffusion language model community for their foundational work
and ongoing contributions to this area of research. 

\bibliography{references}
\bibliographystyle{tmlr}

\appendix
\section{Further Ablations}
\subsection{Noising conditioning time but keeping conditioning data clean}
\label{sec:noise-time-ablation}
\begin{table}[H]
  \centering
  \caption{Effect of noising the conditioning \emph{time} while keeping the
  conditioning \emph{data} clean. $p$ is the probability that a
  conditioning tokens's time value stays clean, i.e remains $t=1$. 95\% bootstrap CIs reported.}
  \label{tab:noise-time-not-data}
  \begin{tabular}{lccc}
    \toprule
    & \multicolumn{3}{c}{Conditioning-time clean prob.\ $p$} \\
    \cmidrule(lr){2-4}
    Difficulty & $0.2$ & $0.5$ & $1.0$ \\
    \midrule
    Easy & $66.95$ \tiny{[64.95, 69.00]} & $69.55$ \tiny{[67.55, 71.60]} & $68.60$ \tiny{[66.60, 70.60]} \\
    Hard & $\phantom{0}4.80$ \tiny{[3.90, 5.75]} & $\phantom{0}5.50$ \tiny{[4.55, 6.50]} & $\phantom{0}3.75$ \tiny{[2.95, 4.60]} \\
    \bottomrule
  \end{tabular}
\end{table}

We ablate whether the improvement in model performance under noised conditioning prompts
stems from the increase in variation of conditioning inputs, hinting at a data augmentation effect,
(i.e model trains on a range of conditioning tokens
$x_t$, not just clean tokens $x_1$) or due to its decrease in reliance on the conditioning prompts for prediction, 
hinting at regularization effect. We only noise the time values of the conditioning tokens
that the model takes as input, but keep the actual conditioning tokens input the same, i.e one-hot vectors. 
Admittedly, we see a confounding effect where the model can learn to identify one-hot vectors as clean, bypassing the 
supplied noised conditioning token time values. In Table~\ref{tab:noise-time-not-data}, we see that
when the model is "told" via the noised time values that the conditioning prompt is noisy, it performs 
simillary on Sudoku Easy and nominally better ($5.50\% \text{ vs } 3.75\%$) on Sudoku Hard.
For significant model improvement, it is necessary to actually noise the conditioning prompt tokens.

\subsection{Separate conditioning time training with $p_{\mathrm{condclean}}$}
\label{sec:seperate-cond-ablation}
We consider an ablation to the seperate conditioning time method (which simply picks an independent time $t'$
to noise the prompt while noising the response at time $t$). We instead have with probability $p_{condclean}$
the prompt remains clean, reducing the training/inference gap. 
Sweeping over $p_{condclean}$ does not lead to significantly different positive results than the original separate conditioning time
training. 
As before, we see that coverage for N-Queens $10\times10$ is still significantly lower than our default shared time 
noising. We leave further investigations into why shared time might have better performance for N-Queens coverage to future work.

\begin{table}[h]
  \centering
  \small
  \setlength{\tabcolsep}{3pt}
  \caption{Conditioning-time-distribution with $p_{condclean}$}
  \label{tab:conditioning-time-distribution}
  \begin{tabular}{lcccccc}
    \toprule
    & \multicolumn{2}{c}{Sudoku} & \multicolumn{2}{c}{N-Queens $n{=}8$}
      & \multicolumn{2}{c}{N-Queens $n{=}10$} \\
    \cmidrule(lr){2-3}\cmidrule(lr){4-5}\cmidrule(lr){6-7}
    $p_\text{condclean}$ & Easy Acc. & Hard Acc. & Accuracy & Coverage
      & Accuracy & Coverage \\
    \midrule
    $0.2$ & $92.38$ & $24.65$ & $97.13$ & $89.11$ & $88.93$ & $52.22$ \\
      & {\scriptsize $[90.35,94.41]$} & {\scriptsize $[16.98,32.32]$}
      & {\scriptsize $[96.61,97.66]$} & {\scriptsize $[88.46,89.76]$}
      & {\scriptsize $[87.71,90.14]$} & {\scriptsize $[51.37,53.06]$} \\
    $0.5$ & $89.66$ & $13.89$ & $97.55$ & $88.62$ & $89.40$ & $52.10$ \\
      & {\scriptsize $[85.62,93.70]$} & {\scriptsize $[9.44,18.34]$}
      & {\scriptsize $[96.80,98.30]$} & {\scriptsize $[88.22,89.02]$}
      & {\scriptsize $[87.15,91.65]$} & {\scriptsize $[51.33,52.86]$} \\
    $1.0$ & $61.32$ & $3.84$ & $97.52$ & $88.54$ & $88.67$ & $51.92$ \\
      & {\scriptsize $[47.22,75.42]$} & {\scriptsize $[0.77,6.91]$}
      & {\scriptsize $[97.03,98.02]$} & {\scriptsize $[87.88,89.20]$}
      & {\scriptsize $[87.54,89.81]$} & {\scriptsize $[51.50,52.35]$} \\
    \bottomrule
  \end{tabular}
\end{table}

\section{Full Results}
\label{sec:full-results}
\begin{table}[!htbp]
  \centering
  \caption{Open dialogue validation NLL. (mean $\pm$ std.\ over $3$ training seeds)}
  \label{tab:valnll-dialogue}
  \begin{tabular}{llcc}
  \toprule
  Train pairs & $p_\text{condclean}$ & Best Val NLL & Checkpoint Step \\
  \midrule
  $191$K   & $1.0$ (paste-in) & $0.6644 {\scriptstyle\ \pm\ 0.0104}$ & $12{,}667$ \\
  $191$K   & $0.5$ & $0.5640 {\scriptstyle\ \pm\ 0.0045}$ & $28{,}667$ \\
  $191$K   & $\mathbf{0.3}$ & $\mathbf{0.5543} {\scriptstyle\ \pm\ 0.0054}$ & $35{,}333$ \\
  \midrule
  $3.38$M & $1.0$ (paste-in) & $0.4425 {\scriptstyle\ \pm\ 0.0073}$ & $97{,}333$ \\
  $3.38$M & $\mathbf{0.5}$ & $\mathbf{0.4133} {\scriptstyle\ \pm\ 0.0035}$ & $138{,}000$ \\
  $3.38$M & $0.3$ & $0.4246 {\scriptstyle\ \pm\ 0.0084}$ & $128{,}667$ \\
  \bottomrule
  \end{tabular}
\end{table}

\begin{table}[!htbp]
  \centering
  \small
  \setlength{\tabcolsep}{4pt}
  \caption{N-Queens accuracy and coverage (\%) under guided sampling at
  $p_\text{condclean}{=}0.2$, $750$ held-out boards $\times\,20$ samples. $\Delta$ is the
  \emph{paired} within-seed change from $w{=}1$. Guidance improves
  accuracy but trades off with coverage.}
  \label{tab:cfg-nqueens-seeds}
  \begin{tabular}{llcccc}
  \toprule
  & & \multicolumn{2}{c}{Accuracy} & \multicolumn{2}{c}{Coverage} \\
  \cmidrule(lr){3-4}\cmidrule(lr){5-6}
  Setting & $w$ & Mean & $\Delta$ & Mean & $\Delta$ \\
  \midrule
  \multirow{4}{*}{$n{=}8$ (length 64)}
    & $1$ & $99.43$ & --- & $\mathbf{92.65}$ & --- \\
    & & {\scriptsize $[99.14,\,99.73]$} & & {\scriptsize $[91.98,\,93.32]$} & \\
    & $5$ & $\mathbf{99.61}$ & $+0.18$ & $90.94$ & $-1.71$ \\
    & & {\scriptsize $[99.37,\,99.85]$} & {\scriptsize $[0.05,\,0.31]$} & {\scriptsize $[90.42,\,91.45]$} & {\scriptsize $[-2.18,\,-1.25]$} \\
  \midrule
  \multirow{4}{*}{$n{=}10$ (length 100)}
    & $1$ & $98.40$ & --- & $\mathbf{73.57}$ & --- \\
    & & {\scriptsize $[98.05,\,98.75]$} & & {\scriptsize $[72.91,\,74.23]$} & \\
    & $2$ & $\mathbf{99.16}$ & $+0.76$ & $67.27$ & $-6.30$ \\
    & & {\scriptsize $[98.97,\,99.35]$} & {\scriptsize $[0.47,\,1.04]$} & {\scriptsize $[66.80,\,67.74]$} & {\scriptsize $[-6.91,\,-5.69]$} \\
  \bottomrule
  \end{tabular}
\end{table}

\subsection{(12x12) N-Queens with disjoint solutions}
\label{app:disjoint-nqueens}
For our ($8\times8$) and ($10\times10$) N-Queens results, we ensure that the training input
boards are disjoint from any hold-out evaluation input boards. However, due to the
total solution space being rather limited at these board sizes ($92$, $724$ solutions respectively),
the set of training solutions fully overlaps with the set of evaluation solutions. 

Therefore we also consider evaluate using hold out solutions which are disjoint from the 
solutions seen in the training. We construct a dataset of ($12\times12$) N-Queens with 
$10{,}000$ training samples, $2{,}000$ validation samples, and a $750$ sample hold out evaluation set, 
ensuring that the training input $\emph{and}$ solutions are disjoint from the hold out evaluation 
input and solutions. Besides the dataset, all training details are kept the same as before
and we report two-sided t-distribution CIs using $5$ seeds. 
As shown in Table \ref*{tab:nqueens12-holdout}, we see that the more heavily noising probability $p=0.2$ performs
the best, with increased accuracy and coverage.

\begin{table}[h]
  \centering
  \caption{N-Queens infill accuracy and coverage (\%) at $n=12$ (length $144$) on a
  input and solution disjoint evaluation set, final checkpoint, $750$ puzzles
  $\times\,20$ samples.}
  \label{tab:nqueens12-holdout}
  \begin{tabular}{lcc}
  \toprule
  & \multicolumn{2}{c}{$n=12$ (length 144)} \\
  \cmidrule(lr){2-3}
  $p_\text{condclean}$ & Accuracy & Coverage \\
  \midrule
  $0.2$ & $\mathbf{39.59}$ & $\mathbf{58.75}$ \\
   & {\scriptsize $[38.31,\,40.88]$} & {\scriptsize $[58.00,\,59.51]$} \\
  $0.5$ & $33.49$ & $51.11$ \\
   & {\scriptsize $[31.62,\,35.36]$} & {\scriptsize $[49.44,\,52.78]$} \\
  $1.0$ (paste-in) & $22.66$ & $38.62$ \\
   & {\scriptsize $[18.50,\,26.82]$} & {\scriptsize $[33.06,\,44.18]$} \\
  \bottomrule
\end{tabular}
\end{table}

\subsection{N-Queens solution diversity at matched validity}
\label{app:diversity}
\begin{figure}[t]
  \centering
  \includegraphics[width=0.8\linewidth]{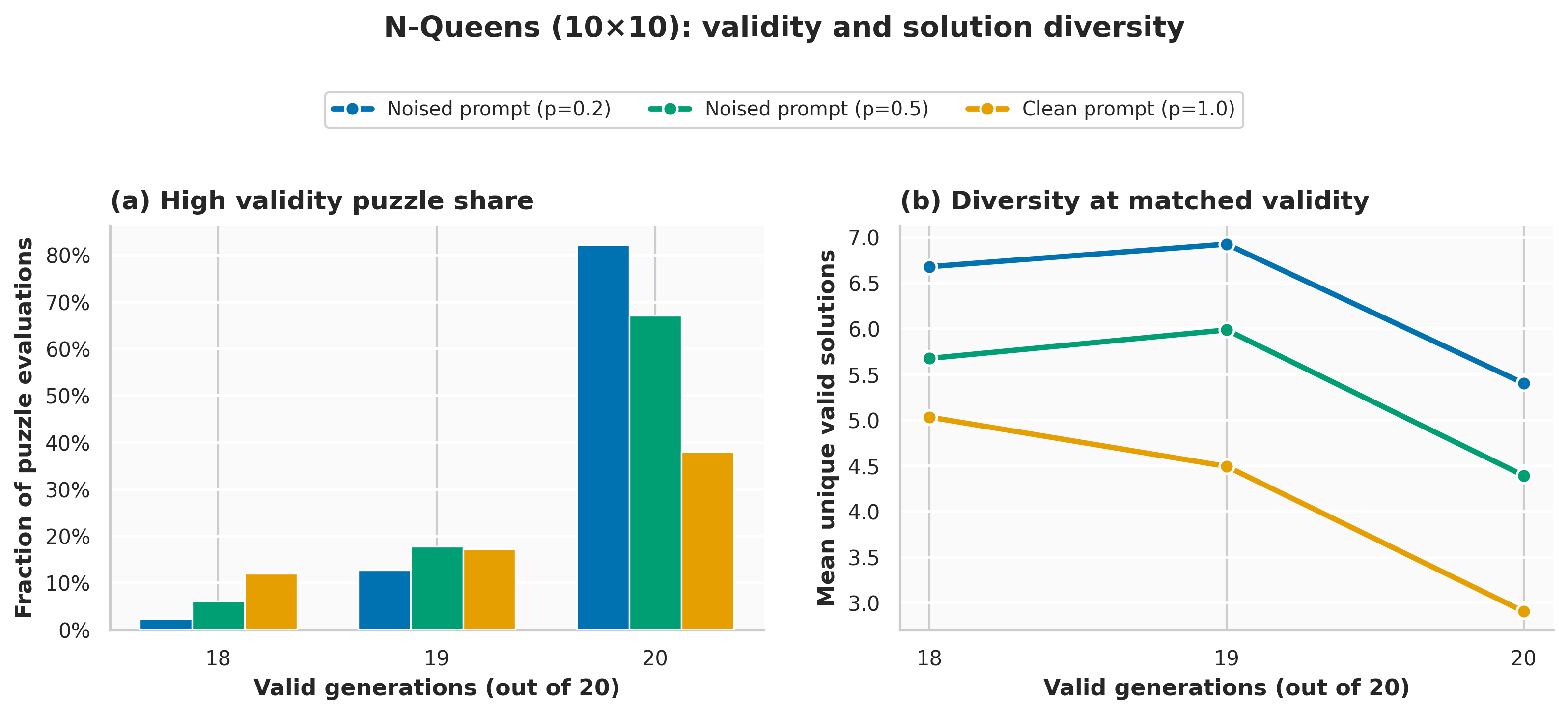}
  \caption{Solution diversity at matched validity for N-Queens $n=10$.}
  \label{fig:nqueens-validity-uniqueness}
\end{figure}

Coverage measures the fraction of total possible solution states generated, which is confounded with 
the model's solution generation quality. While not a significant issue for $n=8$, this is a confounder for $n=10$. Thus
we report the solution diversity at matched validity across the models. As shown in the left half of Figure~\ref*{fig:nqueens-validity-uniqueness},
the vast majority of set-of-20 generations have $18$ or more valid solutions, making up $97.4\%$ of generations
for the $p=0.2$ model and $67.5\%$ of generations for the $p=1.0$ model. On the right we group the set-of-20 generations
by validity counts and report the mean number of unique solutions to measure diversity of generation under matched validity.
We see that the strongly noised $p=0.2$ model upper bounds the default clean prompt $p=1.0$ in terms of solution diversity.  

\subsection{Guidance details}
\label{app:cfg-sweep}
We sweep the guidance weight $w \in \{1, 1.5, 2, 3, 5, 8\}$ on a single training seed for both Sudoku 
and N-Queens experiments. We also consider an affine interpolation in simplex space, i.e exponentiating
the model's log-softmax logits prediction before interpolating and renormalizing the interpolation. However,
we found that interpolating logits to be much more effective so we use this throughout. 
\begin{table}[H]
\centering
\small
\caption{Sudoku solve rate (\%) over the full guidance sweep, $p_\text{condclean}{=}0.2$, using single
seed.}
\label{tab:cfg-sweep-full}
\begin{tabular}{llcccccc}
\toprule
& & \multicolumn{6}{c}{Guidance weight $w$} \\
\cmidrule(lr){3-8}
Setting & Space & $1$ & $1.5$ & $2$ & $3$ & $5$ & $8$ \\
\midrule
\multirow{2}{*}{Easy, $N{=}10{,}000$}
  & logit & $90.90$ & $91.30$ & $91.80$ & $92.20$ & $\mathbf{92.65}$ & $\mathbf{92.65}$ \\
  & prob  & $90.90$ & $91.00$ & $91.15$ & $91.35$ & $91.45$ & $91.55$ \\
\midrule
\multirow{2}{*}{Hard, $N{=}10{,}000$}
  & logit & $28.15$ & $31.60$ & $32.95$ & $34.40$ & $\mathbf{35.35}$ & $34.95$ \\
  & prob  & $28.15$ & $30.05$ & $30.80$ & $31.10$ & $31.70$ & $32.15$ \\
\midrule
\multirow{2}{*}{Easy, $N{=}1{,}000$}
  & logit & $2.05$ & $2.45$ & $3.75$ & $\mathbf{4.60}$ & $4.10$ & $2.65$ \\
  & prob  & $2.05$ & $1.85$ & $1.85$ & $2.50$ & $2.95$ & $3.30$ \\
\midrule
\multirow{2}{*}{Hard, $N{=}1{,}000$}
  & logit & $0.05$ & $0.00$ & $0.05$ & $0.00$ & $0.00$ & $0.00$ \\
  & prob  & $0.05$ & $0.00$ & $0.00$ & $0.00$ & $0.00$ & $0.05$ \\
\bottomrule
\end{tabular}
\end{table}

\begin{table}[H]
  \centering
  \small
  \caption{N-Queens accuracy and coverage (\%) under guided sampling, $p_\text{condclean}{=}0.2$,
  single training seed, $750$ puzzles $\times\,20$ samples. Parenthesised values
  are the change from $w{=}1$.}
  \label{tab:cfg-nqueens}
  \begin{tabular}{lcccc}
  \toprule
  & \multicolumn{2}{c}{$n=8$ (length 64)} & \multicolumn{2}{c}{$n=10$ (length 100)} \\
  \cmidrule(lr){2-3}\cmidrule(lr){4-5}
  $w$ & Accuracy & Coverage & Accuracy & Coverage \\
  \midrule
  $1$   & $99.13$ & $\mathbf{92.65}$ & $97.92$ & $\mathbf{74.38}$ \\
  $1.5$ & $99.18$ {\scriptsize $(+0.05)$} & $91.66$ {\scriptsize $(-0.99)$} & $98.75$ {\scriptsize $(+0.83)$} & $69.13$ {\scriptsize $(-5.25)$} \\
  $2$   & $99.25$ {\scriptsize $(+0.13)$} & $91.26$ {\scriptsize $(-1.39)$} & $\mathbf{98.95}$ {\scriptsize $(+1.03)$} & $67.12$ {\scriptsize $(-7.26)$} \\
  $3$   & $99.29$ {\scriptsize $(+0.17)$} & $90.85$ {\scriptsize $(-1.80)$} & $\mathbf{98.95}$ {\scriptsize $(+1.03)$} & $65.45$ {\scriptsize $(-8.93)$} \\
  $5$   & $\mathbf{99.35}$ {\scriptsize $(+0.23)$} & $91.19$ {\scriptsize $(-1.46)$} & $98.67$ {\scriptsize $(+0.75)$} & $64.22$ {\scriptsize $(-10.16)$} \\
  $8$   & $99.16$ {\scriptsize $(+0.03)$} & $91.05$ {\scriptsize $(-1.60)$} & $97.65$ {\scriptsize $(-0.27)$} & $63.67$ {\scriptsize $(-10.71)$} \\
  \bottomrule
  \end{tabular}
  \end{table}

\section{Full Hyperparameters}
\label{app:hyperparams}
Table~\ref{tab:hparams} lists the shared training configuration.

\begin{table}[h]
\centering
\caption{Shared hyperparameters}
\label{tab:hparams}
\begin{tabular}{lll}
\toprule
Group & Setting & Value \\
\midrule
Model & hidden size / blocks / heads & $512$ / $8$ / $8$ (default), $768$ / $12$ / $12$ (Gigaword 132M) \\
      & cond.\ dim / dropout & $128$ / $0.1$ \\
      & board length & $81$ (Sudoku), $64$/$100$ (N-Queens $n{=}8$/$10$), $80$ (Gigaword) \\
      Algorithm & pred.\ type & $x_1$ \\
      & loss type & flow \\
      & $p_\text{condclean}$ & $\{0.1,0.2,0.3,0.5,1.0\}$ (Sudoku), \\
      &                      & $\{0.2,0.5,1.0\}$ (N-Queens), $\{0.3,0.5,0.8,1.0\}$ (Gigaword), \\
      &                      & $\{0.3,0.5,1.0\}$ (Dialogue) \\
Training & max steps & $100$k (N-Queens), $40$k (Sudoku), $60$k (Gigaword), $150$k (Dialogue) \\
         & global batch size & $256$ \\
         & precision / EMA & bf16-mixed / $0.9999$ \\
         & grad clip & $1.0$ \\
Optimizer & class & AdamW \\
          & lr / $(\beta_1,\beta_2)$ & $3\!\times\!10^{-4}$ / $(0.9,0.999)$ \\
          & warmup & $2{,}500$ steps (constant after) \\
Sampling & solver & Euler \\
         & steps & $128$ \\
Hardware & accelerator / GPU & cuda, 1 device / NVIDIA L40S \\
\bottomrule
\end{tabular}
\end{table}

\end{document}

%% file: math_commands.tex
\usepackage{amsmath,amsfonts,bm}

\def\eqref#1{equation~\ref{#1}}

\def\1{\bm{1}}

\DeclareMathAlphabet{\mathsfit}{\encodingdefault}{\sfdefault}{m}{sl}
\SetMathAlphabet{\mathsfit}{bold}{\encodingdefault}{\sfdefault}{bx}{n}

